\documentclass[journal]{IEEEtran}

\usepackage[style=ieee]{biblatex} 
\bibliography{example_bib.bib}    
\usepackage{hyperref}
\usepackage{amsmath}
 \usepackage{multirow}

\usepackage{url}
\usepackage{xcolor}
\usepackage{graphicx}  
\usepackage{caption}
\usepackage{float}  

\usepackage{titlesec}

\begin{document}

\title{ \LARGE Experiments in Autonomous Driving Through Imitation Learning \\ }

\author{Michael Muratov,
        Abdulwasay Mehar,
        Wan Song Lee \\
        Michael Szpakowicz,
        Ose Edmond Umolu\\
        Joshua Mazariegos Bobadilla,
        Ali Kuwajerwala
        }

\markboth{\MakeSentenceCase{{CSC}493 {R}eport, {U}niversity of {T}oronto, {A}pril 2020}}
{Shell \MakeLowercase{\textit{et al.}}: Bare Demo of IEEEtran.cls for IEEE Journals}

\maketitle

\begin{abstract}This report demonstrates several methods used to make a self-driving vehicle using a supervised learning algorithm and a forward-facing RGBD camera. The project originally involved research in creating an adversarial attack on the vehicle's model, but due to difficulties with the initial training of the car, the plans were discarded in favor of completing the imitation learning portion of the project. Many approaches were explored, but due to challenges introduced by an unbalanced data set, the approaches had limited effectiveness. 
\end{abstract}


\section{Introduction}

\IEEEPARstart{S}{elf} driving cars are fully dependent on how they process data from their array of on-board sensors, such as RGB and depth cameras, LIDARs, and IMUs. The goal of this project was to use these same types of sensors to produce effective self driving behaviour in an indoor environment. The methods in this report were implemented on the ROS-based MIT Racecar platform with a Hokuyo Rangefinder and a Intel RealSense Depth Camera. The next steps in this project would be to undermine the car's decision-making model using a plausible black box adversarial attack. Unfortunately, due to the closure of the university, there were major setbacks to the project which impacted both the development of the self driving network and performing adversarial attacks on it.

\section{PID Control}
The project began with the implementation of a PID-controlled wall-following algorithm for automated training and label collection. The code for the algorithm is currently in a private repository due to the material being very similar to the course material of CSC477.

\subsection{Wall Following}
The car's LIDAR sensor measured distance markers in front and to the sides of the car, keeping track of the closest marker to determine its distance away from the wall. It used the PID controller to stay at the set distance away from the wall at any given time.

\subsection{Parameter Tuning}
The Ziegler-Nichols method was used to tune the PID controller's parameters. The result produced a car which did a good job of staying in the middle of the hallway.

\subsection{Issues with PID}
Although effective, the PID algorithm was too erratic, and produced labels which changed drastically between frames. The use of PID-generated labels for data collection was abandoned since they were noisy and suspected to be very difficult to train on. 

\section{Collecting Data}
Collecting data for the machine learning data set was very straightforward due to the advanced sensors and processing unit present on the vehicle. After initial setup, the data was collected automatically as a human expert drove the car around the given course. Although the data collection process was simple, the initial set up for it was very complex. Due to this inherent complexity, there was not enough data collected at the beginning of the project. Sadly, this only became apparent after the campus closed, and collecting more data was no longer an option. The data collection script can be found at:\\
\textcolor{blue}{\url{https://github.com/Autonomous-Robotics-UTM/racecar\_ws/blob/master/src/realsense-ros-2.2.8/realsense2\_camera/scripts/record\_data.py}}

The instructions for initiating the data collection can be found in the repository's README file. 

\subsection{RGB Images}

Gathering RGB images was straightforward with the Intel Realsense Camera ROS SDK provided by Intel's Github repository: 
\textcolor{blue}{\url{https://github.com/IntelRealSense/realsense-ros}}

Using the camera, the car captured ROS Image objects as they were created, converted them into OpenCV arrays, and saved them to a designated folder as jpg files.

\begin{figure}[H]
\centering
\includegraphics[width=0.3\textwidth]{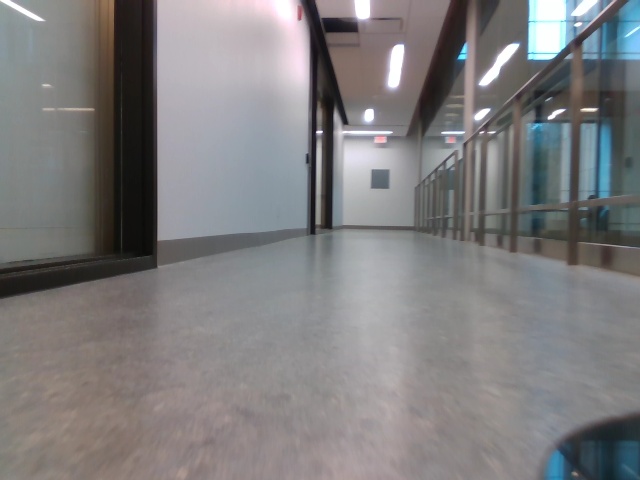}
\caption{Collected Image file}
\end{figure}

\subsection{Labels}
The labels were also collected in real time based on the expert's joystick inputs. The labels were then synchronized to their corresponding images using the ApproximateTimeSynchronizer message filter, and they were only saved if they had a corresponding image to go along with. These labels were saved to a CSV file where they could be easily retrieved during training. The following table shows a sample CSV file filled with these labels.

\begin{center}
 \begin{tabular}{||c c||} 
 \hline
 ID & Angle\\ [0.5ex] 
 \hline\hline
 0 & 0\\ 
 \hline
 1 & -0.15\\
 \hline
 2 & 0\\
 \hline
 3 & 0.08\\
 \hline
 4 & 0.21\\
 \hline
 ... & ...\\ [1ex] 
 \hline
\end{tabular}
\end{center}

\subsection{Depth Information}

Capturing the depth information proved to be more difficult because the camera would store the information as a point cloud. This made it difficult to combine the previously captured RGB image with the depth information. This was fixed by setting the align\_depth:=true flag when running the rs\_camera node to produce a RGBD ROS topic.

\begin{figure}[H]
\centering
\includegraphics[width=0.3\textwidth]{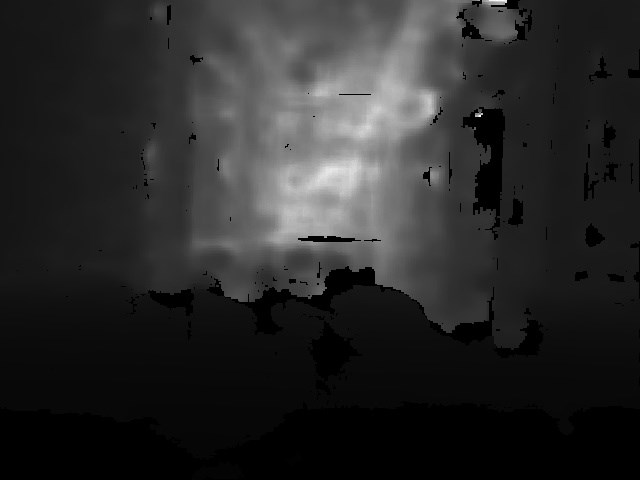}
\caption{Depth Image from the RealSense Camera}
\end{figure}

\subsection{Issues with Data Collection}
The main problem of the data collection was the gross over-representation of the label corresponding to moving straight. Not only were there many images that corresponded to moving straight, there were also many labels which corresponded to small corrective movements. These labels may have confused our network when deciding between images which showed small corrections, and those which showed straight forward motion. These issues could have easily been overcome by collecting more data, however, this was not realized until much later. At this point in the project, a lot of time was spent in vain attempting to "fix" the network before realizing the data collected was imbalanced.

\section{Computing the Loss}
The self-driving car's steering angle was controlled by a neural network while its wheels spun forward at a constant velocity. Below are the attempts at creating a working driving network.
  
\subsection{Regression}
The first approach at steering a car with a neural network was to compute the mean squared error between our network's prediction and the target label for values between -1 and 1. If the network performed well at predicting the labels, the distance between each prediction and corresponding label would be small. When using this approach, the network would consistently produce a label that was closest to the 0 label (straight) which was the highest peak in our data set, and slightly leaning towards -1 (right), where the tail of our data set was found. This caused poor performance because the network would issue the same steering command for any input image. This effect can be seen in the image below. The correct path is drawn in green, and the model's path is drawn in blue. The network is always choosing to move straight and slightly to the right, because the car is biased towards the left and the data gathered is compensating for that. This results in the car moving in circles.

\begin{figure}[H]
\centering
\includegraphics[width=0.3\textwidth]{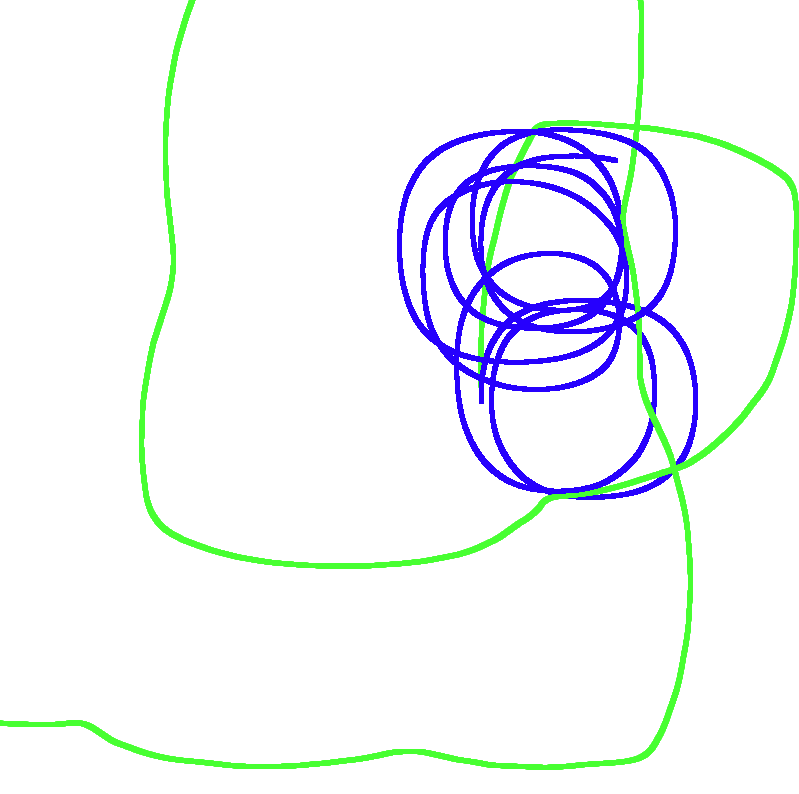}
\caption{Motion of the labels and the predictions}
\end{figure}

The result of the regression approach can be seen implemented on the web here:\\
\textcolor{blue}{\url{https://cs.utm.utoronto.ca/~muratovm/Car/index.html}}
The network was converted into an ONNX model and run on the browser with the ONNX library. It is preferred to use a good computer to view the network in action because JavaScript is not an efficient language.

\subsection{Categorical}
Another approach was breaking up the predictions into distinct categories (or classes) in the range of -1 to 1 and then training the network to classify images into the right categories using cross entropy loss. Still, this approach was prone to the same pitfall as the regression approach - the model would lean towards always categorizing images as moving straight, because that was the category of images which appeared most often.

\begin{figure}[H]
\centering
\includegraphics[width=0.4\textwidth]{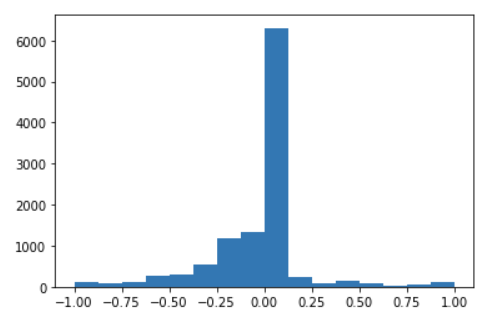}
\caption{Distribution of the Labels}
\end{figure}

\subsection{Inverse Weighting}
To combat the above problem of the network sticking to predicting the most common category of the data set, each category was weighed inversely proportionally to their rate of occurrence. This approach seemed to work well, producing varied results, until the network was able to over fit onto the data set again and produce the same issues as in the unmodified categorical case. 

\subsection{Inverse Gaussian Weighting}
The idea behind this approach was to force the network to produce Gaussian curves with the peak of the curve at the position of the correct category. This would be achieved by weighting the categories which were further away from the correct more than those which were closer. The weighting would follow an inverse Gaussian distribution. Due to poor organization, this approach was not fully implemented and untested. It is unclear whether this approach would have been successful based on the issues with the data set and the outcomes of the previous attempts.

\section{Augmenting the Data set}

\subsection{Omitting Images}
Omitting some of the the images labeled straight worked well for the simple CNN model. The model produced more varied results than when using the entire training set. The issue was that the model was not sophisticated enough to learn actual correct labels; even though the results were diverse, they were not accurate. This approach also did not work for implementations where several images were taken together to form a series. Omitting images would break the sequential nature of the video.
\subsection{Reflection}
Reflecting the images was an easy way to double the amount of data we had and removed the bias to move the the right instead of to the left. This approach however also doubled the number of straight labeled images, which then had to be omitted.

\begin{figure}[H]
\centering
\includegraphics[width=0.45\textwidth]{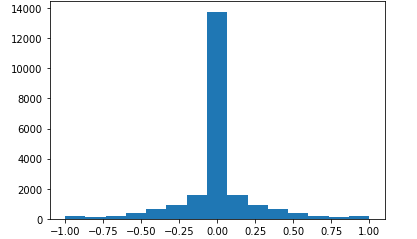}
\caption{Distribution of the Labels with Reflection Augmentation}
\end{figure}

\begin{figure}[H]
\centering
\includegraphics[width=0.4\textwidth]{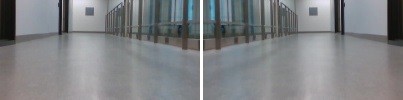}
\caption{Vertically Flipped Image}
\end{figure}

\subsection{Gaussian Noise}
Introducing Gaussian noise increased the number of images that could be used in training the pure CNN model. This approach was tested with the built-in ResNet model with transfer learning and it was found that it produced no effect on our training accuracy. 

\begin{figure}[H]
\centering
\includegraphics[width=0.3\textwidth]{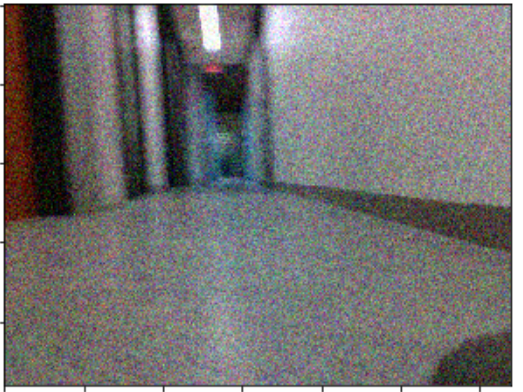}
\caption{Image with exaggerated Gaussian Noise}
\end{figure}

\subsection{Colour Schemes}
Choosing colour schemes other than RGB was explored in hopes that the network could pick up on the colour information to make better decisions. Using HSV, LAB, YCB and a custom variant of HSV which decomposed H into X and Y, did not yield improved results compared to standard RGB. 

\begin{figure}[H]
\centering
\includegraphics[width=0.5\textwidth]{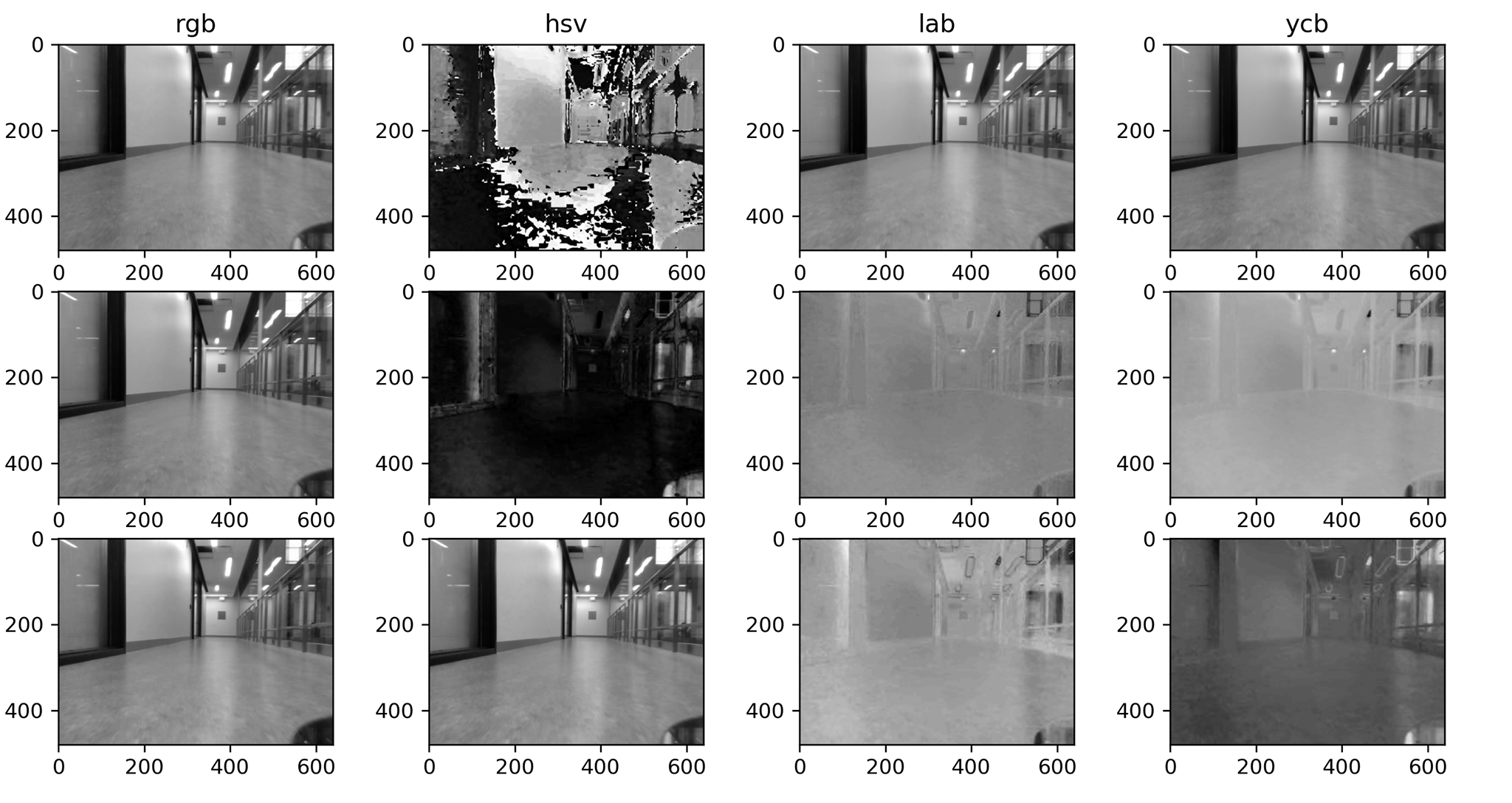}
\caption{Attempted colour Formats}
\end{figure}

\subsection{Cropping and Scaling}
Cropping and Scaling the image was very helpful in reducing the amount of time the network spent training and reducing the amount of unnecessary information it would have to process. We took original images with dimensions 480x640, cropped a portion of dimensions 220 x 400 and interpolated it down to 100x200. Theoretically, this approach undoubtedly was an improvement since it removed unnecessary data and increased generalization. However, due to the failures in our data collection, we were not able to see tangible improvement.

\begin{figure}[H]
\centering
\includegraphics[width=0.5\textwidth]{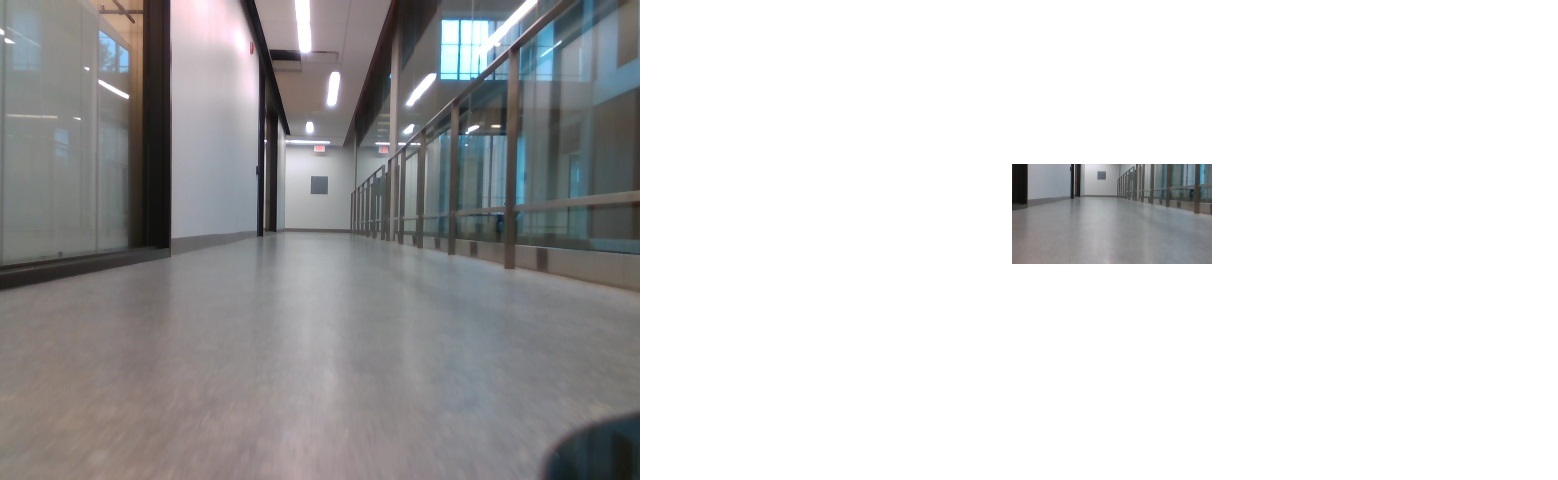}
\caption{Cropping and Scaling an Image}
\end{figure}

\subsection{PyTorch Image Segmentation}
PyTorch has a built-in network for segmenting images. The modified training set was passed through this network, and its results were passed as as inputs into the predictor network. The segmentation network was only able to segment the floor from the rest of the image and not enough testing was done to deem whether it was beneficial or not.  

\begin{figure}[H]
\centering
\includegraphics[width=0.5\textwidth]{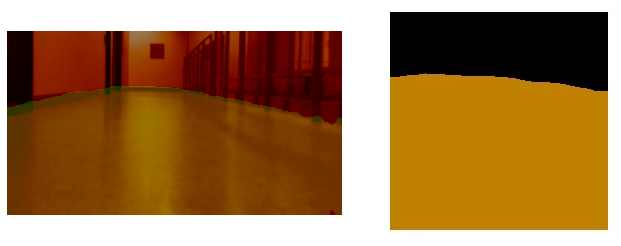}
\caption{PyTorch Floor Segmentation}
\end{figure}

\subsection{K-Means Image Segmentation}

When deciding on what features of the image to include, reducing the number of colours in the image helped with conventional computer vision techniques. This was done by running multiple passes of OpenCV's built in K-Means function on a vectorized image. The result of the process was an image with as many distinct colours as the passed-in value of K. Below are three different images where K = 2, 5 and 9 in ascending order. Having less colours reduces the image into simple shapes, although as seen with k= 5 and K = 9, there are issues with the lighting bouncing off the floor, creating shapes which are not really there. For this reason we chose K=2 when using Computer Vision techniques. 

\begin{figure}[H]
\centering
\includegraphics[width=0.3\textwidth]{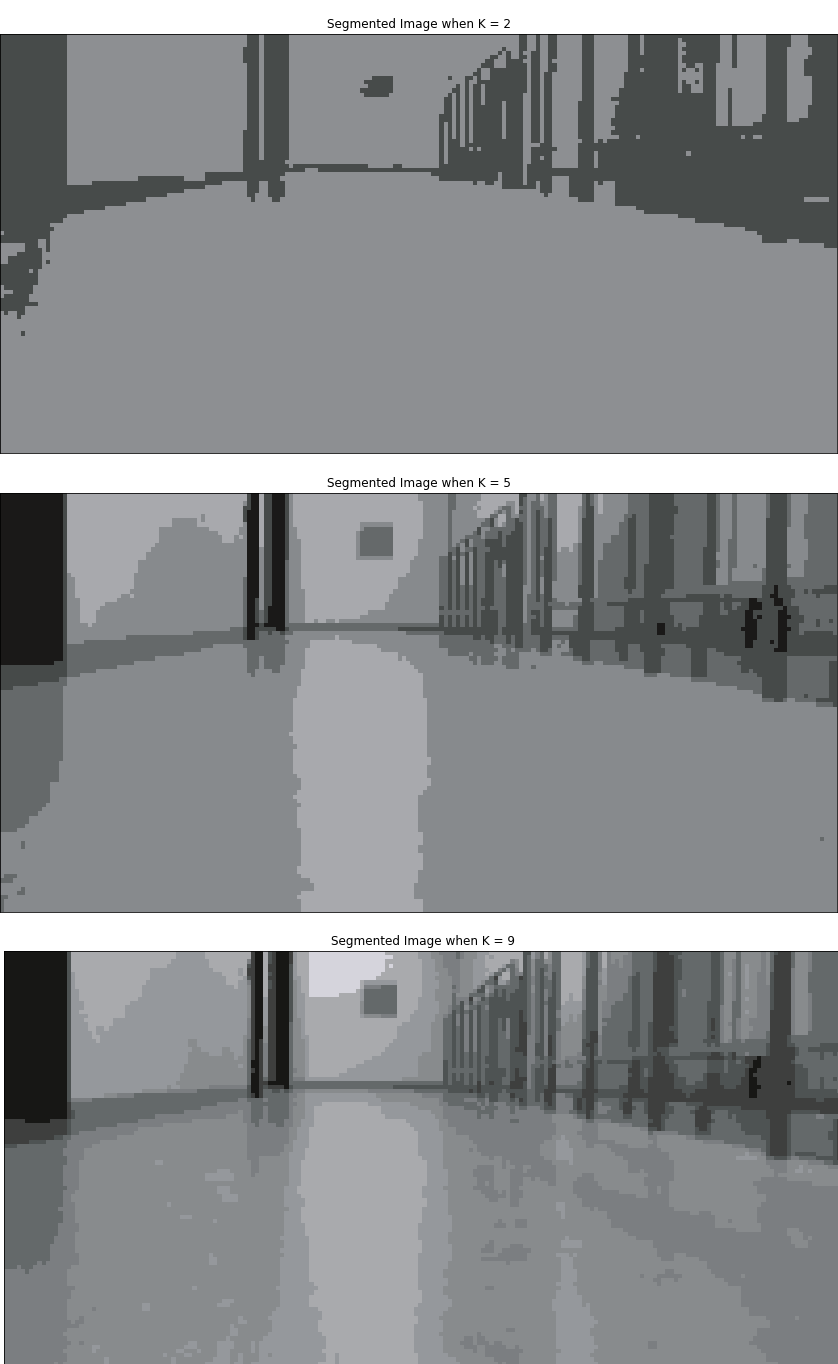}
\caption{K-Means Image Segmentation with K=2, K=5 and K=9}
\end{figure}

\subsection{Hough Lines}
The Computer Vision approach we decided to take was the Hough Line Method. The method takes vertices of interest and draws lines where edges are detected. As seen from the image below, running Hough Lines on the K-Means Segmented image produces better results than when run on the standard RGB image. This is because the segmentation reduces complexity and allows us to see the most prominent edges. There was not enough time to finish our experimentation with Hough Lines however, and it was left as an exercise for the future. 

\begin{figure}[H]
\centering
\includegraphics[width=0.5\textwidth]{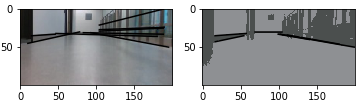}
\caption{Hough Lines with and without K-Means Segmentation}
\end{figure}

\section{Attempted Architectures}
\subsection{Convolutional Neural Network}
The Convolutional Neural Network tested had 3 convolutional layers with 16, 32, and 64 filters respectively, followed by 3 linear layers which gradually decreased in size and produced the prediction in the final layer. The network was kept constant whether the output was a label prediction or category. This network was not producing any improvement, and based on the following results, it is first speculated that it was not that model that was at fault but the data we were attempting to teach it. This model was able to overfit on a small amount of data, but when attempting to generalize on the entire data set, it produced the results seen below.

\begin{figure}[H]
\centering
\includegraphics[width=0.5\textwidth]{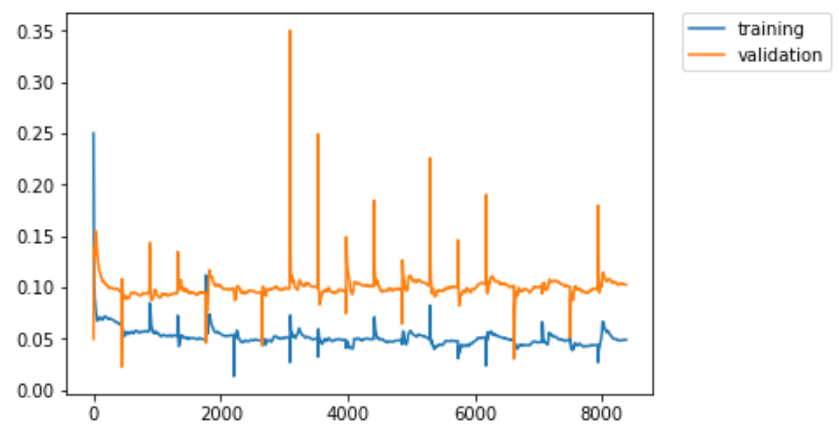}
\caption{CNN Loss curves for Training and Validation}
\end{figure}

The labels were also scaled in an attempt to provide a stronger gradient signal through the network. The results were somewhat different to those of past attempts, but in the end, the network was no better at predicting the correct labels than before.

\begin{figure}[H]
\centering
\includegraphics[width=0.5\textwidth]{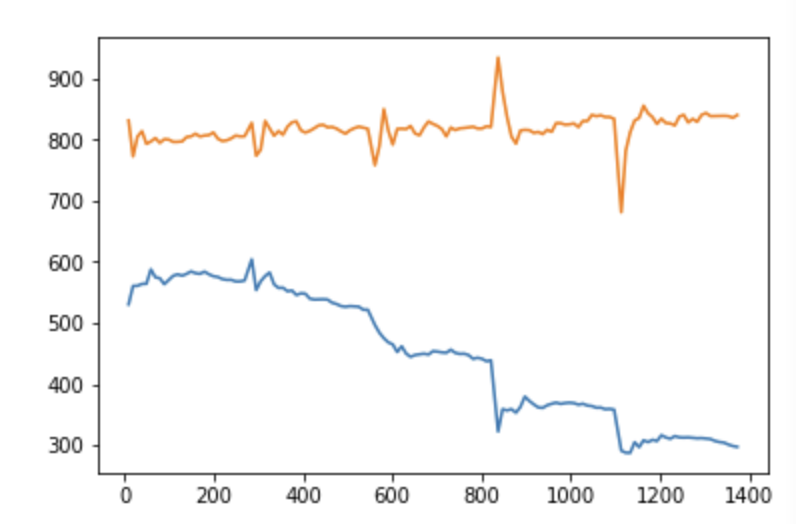}
\caption{CNN Loss curves for scaled labels}
\end{figure}

The activations for the convolutional layers do seem reasonable and can be seen below. The last layer convolutions are picking up on the edges of the image but it does not seem to be enough to predict the precise labels. 

\begin{figure}[H]
\centering
\includegraphics[width=0.5\textwidth]{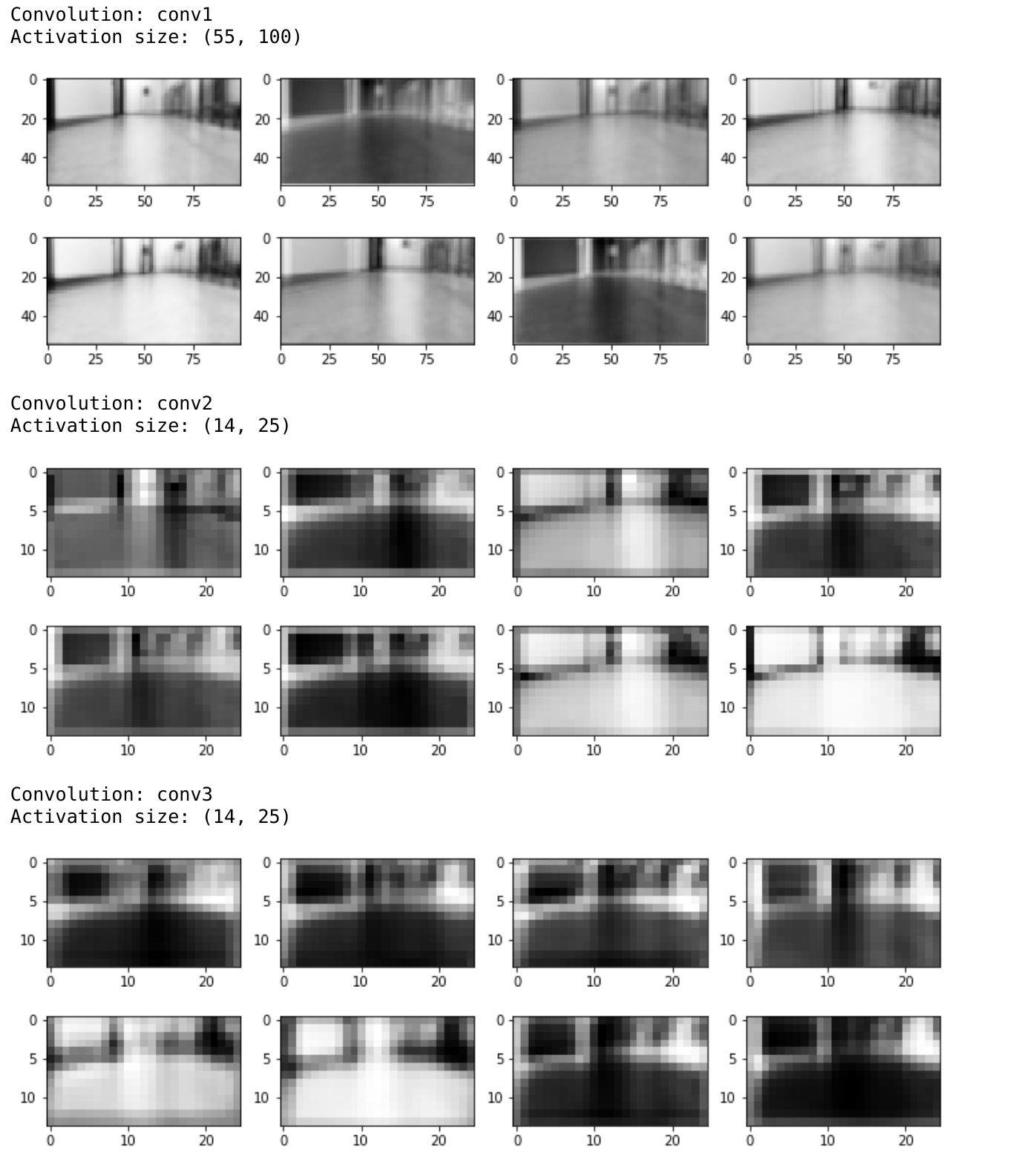}
\caption{CNN layer activations}
\end{figure}

\subsection{ResNet}

There was an attempt at transfer learning by taking a pre-trained ResNet model and replacing its last layer with our own linear network. The results were unfortunately very similar to the previous CNN model along with the expense of running a larger network. It is possible that training the entire ResNet from scratch could produce better results because of the extreme specialization of the task, but trying to do so took too much time with no noticeable improvements.  

\begin{figure}[H]
\centering
\includegraphics[width=0.5\textwidth]{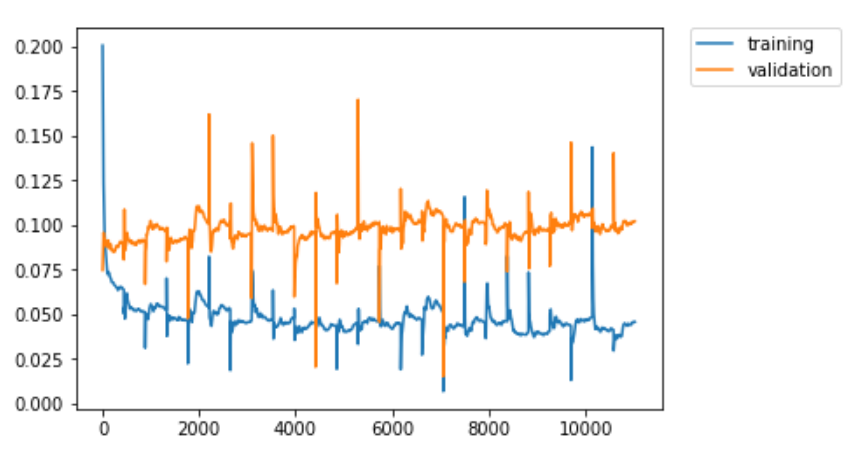}
\caption{Hough Lines with and without K-Means Segmentation}
\end{figure}

\subsection{Nvidia Network}
Since Nvidia already had a successful attempt at modeling images with steering commands, their architecture was trained on the hallway dataset. Their network was comprised of 5 convolution layers and 3 fully-connected linear layers, in which it took in a normalized input (made out of left, center, right camera images). The combination of the three images allowed image shifting augmentation, enabling the network to recover from bad orientation. Since the car had a single camera to its disposal, image shifting would end up losing information. Instead of using regression (as used in the Nvidia network) the network was switched to classification as it would allow us to predict in ranges. This method however also gave a similar over-fitting result as the previous attempts.

\begin{figure}[H]
\centering
\includegraphics[width=0.4\textwidth]{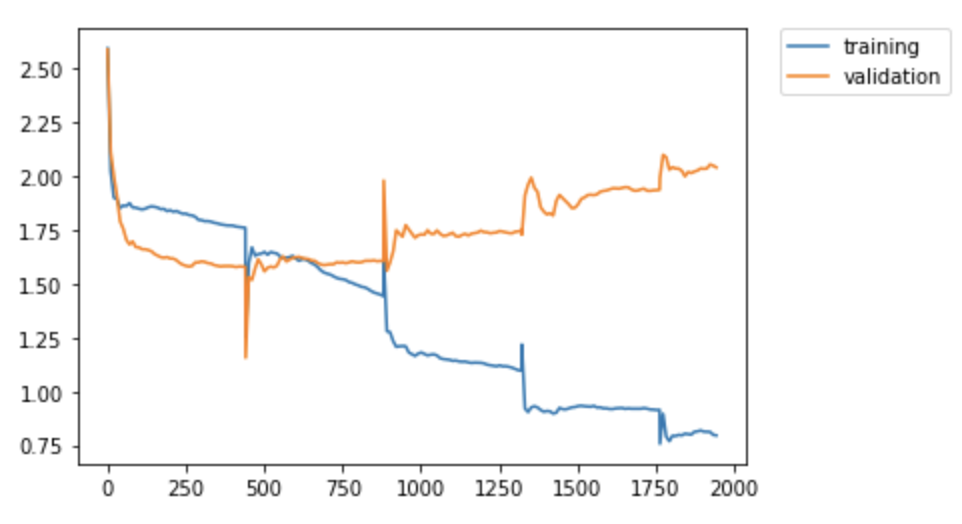}
\caption{Nvidia classification network output}
\end{figure}
This issue indicated an ill-conditioned curve in our loss function, since the model attempted to predict the most common category (i.e going-straight). To tackle this issue, Xavier weight initialization was added for each layer of the network. This method aimed to initialize the network in such as a way that it maintained the output variance for each layer. Doing so kept the error signal from vanishing/exploding during backpropagation, which caused the weights to move towards a lower minimum.

\begin{figure}[H]
\centering
\includegraphics[width=0.4\textwidth]{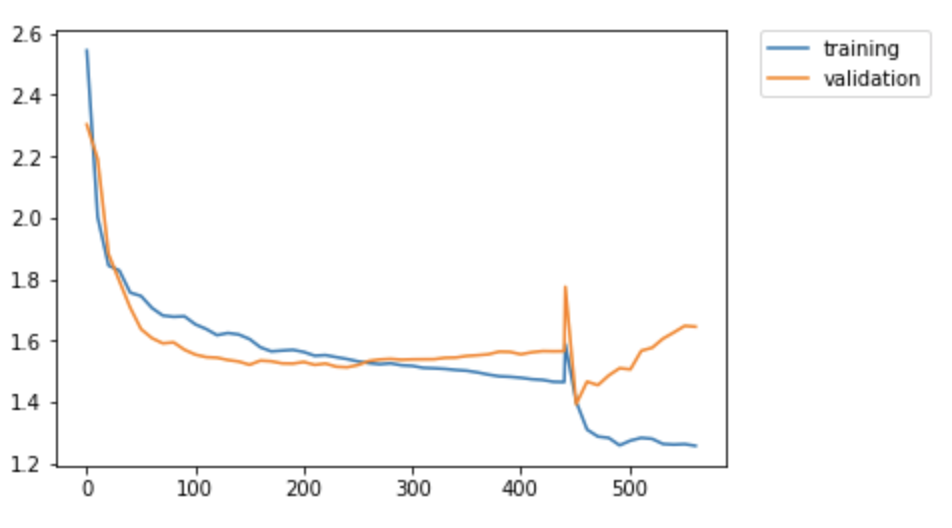}
\caption{Nvidia network with Xavier initialization}
\end{figure}
As it can be seen in the Fig-18, the network showed stable training in comparison to the network without Xavier-initialization. However; the over-fitting issue appears eventually here as well.

\subsection{Auto Encoder + RNN}
The idea behind this approach was to use a Recurrent Neural Network to capture the sequential nature of the series of images to make predictions about the future. Since using the entire image as the input to the network would be too large, we chose to train an Autoencoder to compress the images to an embedding space of 10 variables. Once the Autoencoder was trained, it produced decoded images like the one seen in Fig. 19. 

\begin{figure}[H]
\centering
\includegraphics[width=0.4\textwidth]{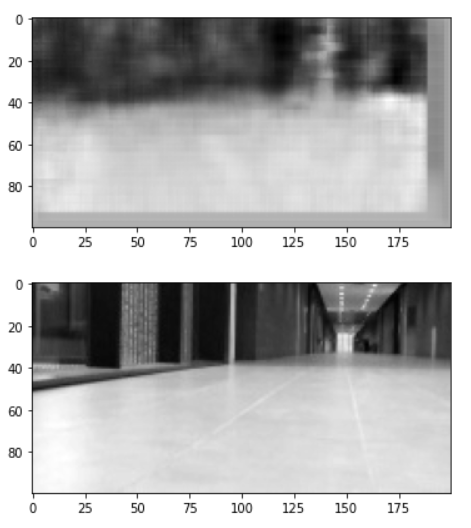}
\caption{Autoencoder output for an embedding size of 10 variables}
\end{figure}
The embedded images were strung together into sequences of 3 images and passed into a GRU block in batches. The GRU block created its own hidden representation of size 10 which fed into a linear layer to convert the representation into a prediction. The resulting network did learn up to a point where it was similar in performance to the past approaches but unfortunately it was not able to provide accurate results. This approach was more complicated than anything tested before, and it is possible that given a more balanced dataset and a larger Autoencoder embedding size, that it would be able to generalize better than the previous approaches due to its ability to model sequential information.

\section{Adversarial Attack}

The original goal of the project was to create a black box adversarial attack that would misdirect the trained model into generating incorrect steering directions. A number of initial materials were created to make that possible, but due to the difficulties faced with creating a functional driving network, the adversarial part of the project was never fully began. Fig. 20 contains an image of a function built to take an image and generate noise on the masked surface. The image with noise would be perspective shifted and placed into the car's field of view. The idea was to simulate the malicious noise on the car and to tune it further, making it influence the cars decisions. The approach would be extended by generating the noise on a TV screen instead of in simulation.

\begin{figure}[H]
\centering
\includegraphics[width=0.5\textwidth]{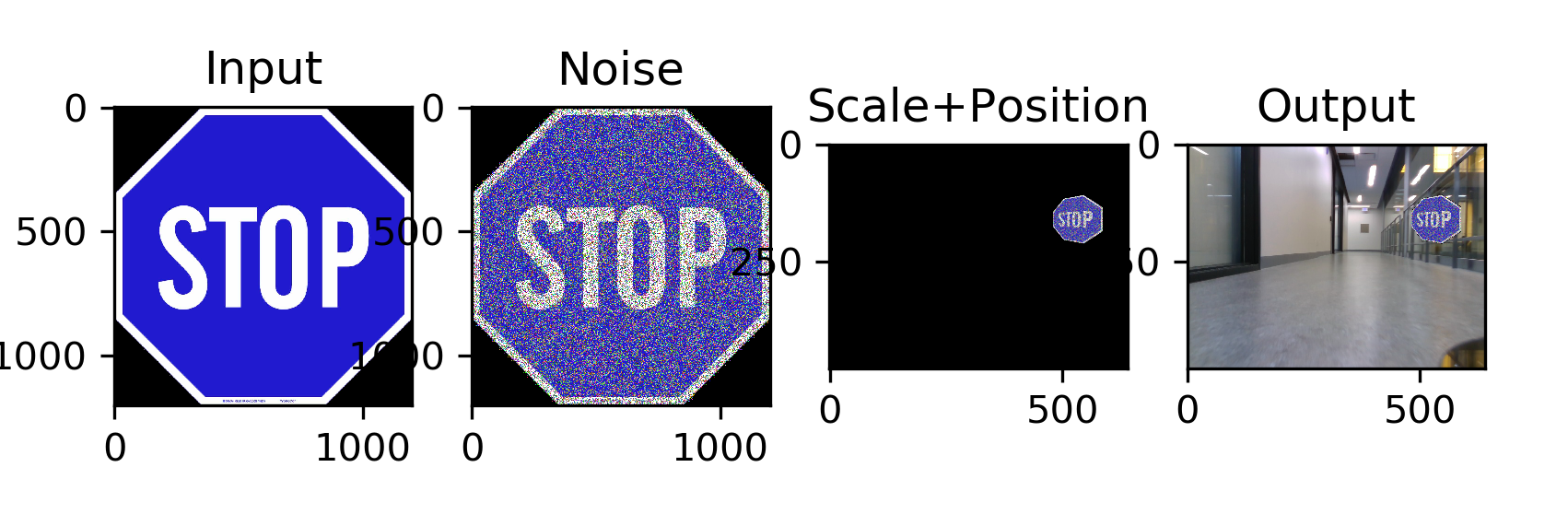}
\caption{Noisy data introduced into the car scene}
\end{figure}
\section{Additional Work}

\subsection{Flask Server with DASH}

A Flask server was created on the car for the purpose of plotting its motions as the car drove around the hall. Commands could also be easily sent to the car remotely through a command terminal set up on a locally hosted website. The terminal could also be run directly on the car to execute the same commands. The eventual plan was to build out the interface to allow anyone not skilled with operating the vehicle to quickly familiarize themselves with all of the functions through the GUI display.

In order to plot the live graph of the car's motion the DASH library was used, which has built in functionality for sending information to and from objects on the website. DASH's timer object was used to update the graph based on its CSV file, which would be updated through a POST request whenever the car collected data. To more easily see the changes in motion, the DASH graphs were built to display previous CSV files that were stored on the car itself, plotted next to or on top of the live graph to see a direct comparison of the motion. The graph could be viewed in depth with built-in DASH graph tools such as zooming in and looking at distinct points. Custom tools were created such as a pause button to stop receiving live data, the ability to limit how many points are displayed on the graph, and displaying the number of captured images while the car was collecting data.

\begin{figure}[H]
\centering
\includegraphics[width=0.5\textwidth]{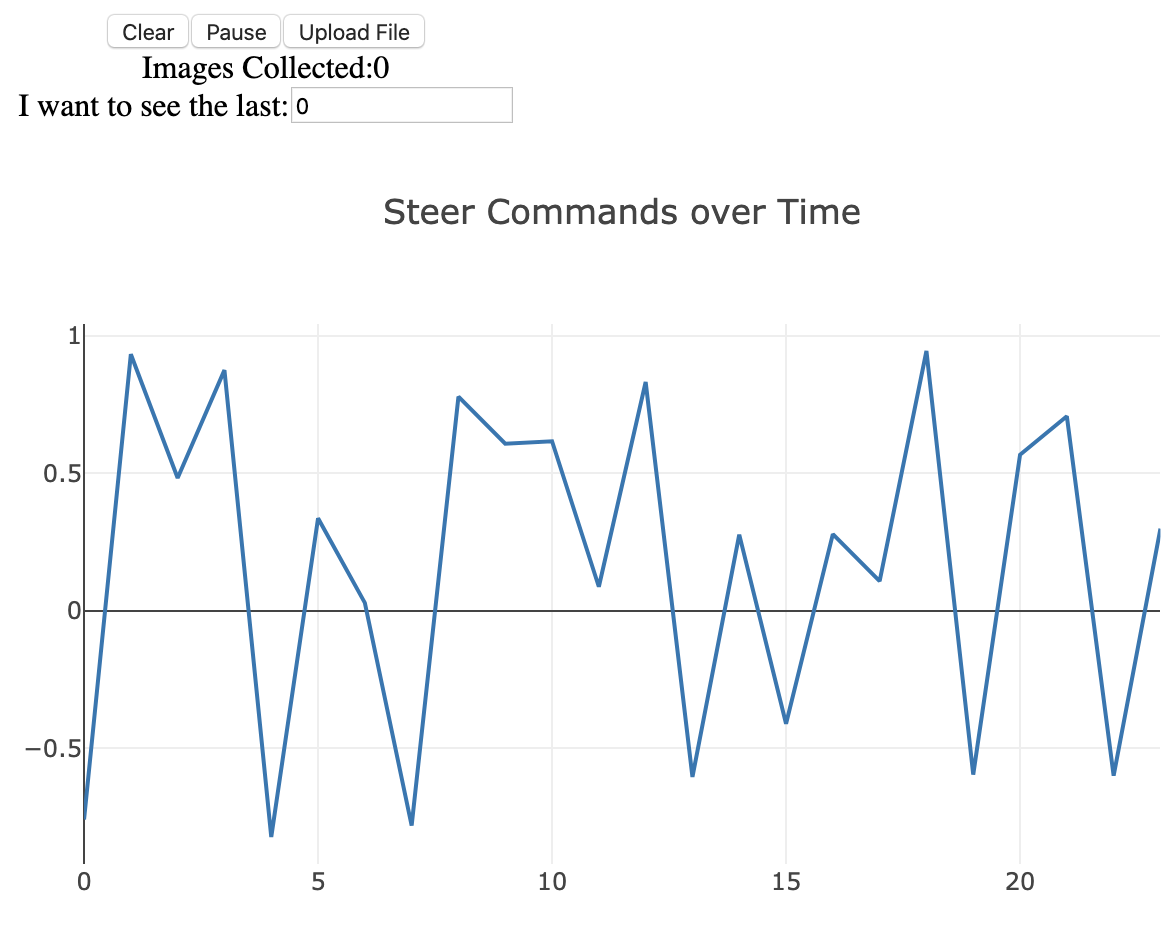}
\caption{Example Live Graph with Tools}
\end{figure}

The console was re-purposed from \textcolor{blue}{\url{https://github.com/ethanhuang0526/javascript-iframe-demo}} and was also hosted on the Flask server. The console also works for sending POST requests where the page itself creates a JSON with the command and then executes it on the car, after which the car's response is returned to the terminal screen. 

\begin{figure}[H]
\centering
\includegraphics[width=0.5\textwidth]{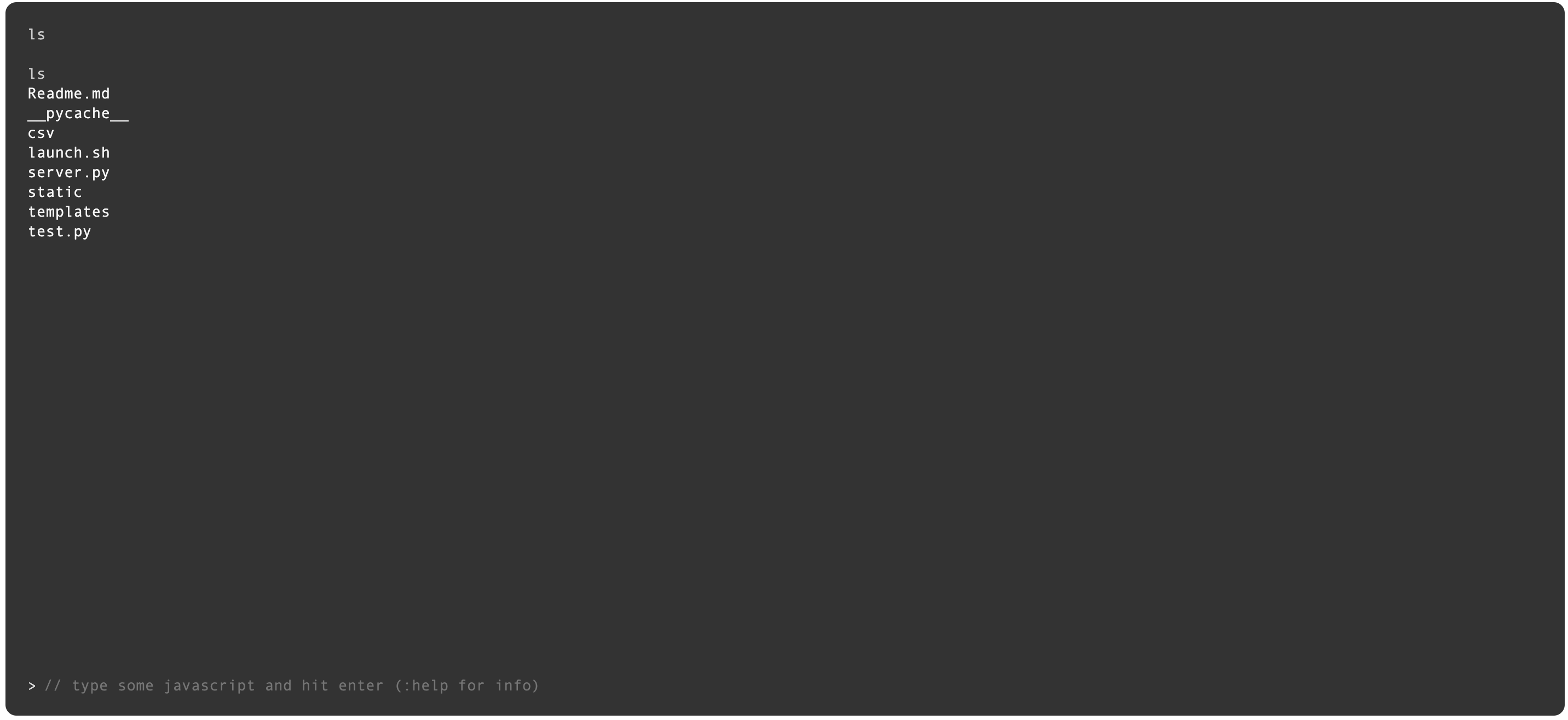}
\caption{Console with an Example 'ls'}
\end{figure}

\section{Conclusion}

We believe that most of the techniques attempted had great promise of success. Despite this, the approaches fell short of accomplishing the goal set out in the beginning of the semester. Nevertheless, the project gave us a chance to work extensively with PyTorch and as a result we were able to develop a range of approaches and test them on real vehicles. We hope that our created materials and results will be helpful to future students attempting to solve the autonomous driving problem.\\ 
All of the technical aspects of this projects can be found at \url{https://github.com/Autonomous-Robotics-UTM}

\section{Acknowledgements}

We would like to thank Professor Florian Shkurti for his guidance during the development of this project, for giving us the opportunity to work on real autonomous vehicles and for having the freedom of taking the research into our own direction. We would also like to thank grad students Dhruv Sharma and Reinhard Grassmann for their eagerness to help us with all of our questions.


\section*{References}

[1]  Y. Li and S. Birchfield. “Image-Based Segmentation of Indoor Corridor Floors for a Mobile Robot”. 1-7, 2018.\\ [0.1in]

[2]  Morgulis, N., Kreines, A., Mendelowitz, S., \& Weisglass, W. (2019, June 30). Fooling a Real Car with Adversarial Traffic Signs. Retrieved from \url{https://arxiv.org/abs/1907.00374}\\ [0.1in]

[3] Kocic, J., Jovicic, N., \& Drndarevic, V. (2019, May 3). An End\-to\-End Deep Neural Network for Autonomous Driving Retrieved from \url{https://www.researchgate.net/publication/332849707\_An\_End-to-End\_Deep\_Neural\_Network\_for\_Autonomous\_Driving\_Designed\_for}\\ [0.1in]

[4] Pomerleau, D. A. (n.d.). Neural Network Perception For Mobile Robot Guidance. Retrieved from \url{https://link.springer.com/book/10.1007/978-1-4615-3192-0}\\ [0.1in]

[5] Bojarski, M., Firner, B., Jackel, L., Muller, U., Zieba, K., Flepp, B., \& Testa, D. D. (2018, April 25). End-to-End Deep Learning for Self-Driving Cars. Retrieved from \url{https://devblogs.nvidia.com/deep-learning-self-driving-cars/}\\ [0.1in]

\end{document}